\documentclass[letterpaper]{article} 
\usepackage{aaai24}  
\usepackage{times}  
\usepackage{helvet}  
\usepackage{courier}  
\usepackage[hyphens]{url}  
\usepackage{graphicx} 
\usepackage{natbib}  
\usepackage{caption} 
\usepackage{algorithm}
\usepackage{algorithmic}
\usepackage{amsmath,amsfonts}
\usepackage{array}

\usepackage{threeparttable}
\usepackage{textcomp}
\usepackage{url}
\usepackage{verbatim}
\usepackage{graphicx}
\usepackage{cite}
\usepackage{color}
\usepackage{colortbl}
\usepackage{xcolor}
\usepackage{multirow}
\usepackage{bbding}
\usepackage{caption}
\usepackage{footnote}
\usepackage{shadowtext}
\usepackage{booktabs}
\usepackage{amssymb}
\usepackage{subfig} 
\usepackage{subfloat}
\usepackage{multicol}
\usepackage{listings}
\usepackage{xcolor}
\nocopyright
\usepackage{newfloat}
\usepackage{listings}
\DeclareCaptionStyle{ruled}{labelfont=normalfont,labelsep=colon,strut=off} 
\floatstyle{ruled}
\newfloat{listing}{tb}{lst}{}
\floatname{listing}{Listing}
\title{SCD-Net: Spatiotemporal Clues Disentanglement Network for Self-supervised Skeleton-based Action Recognition
}
\author{
    Cong~Wu\textsuperscript{\rm 1,\rm2},
    Xiao-Jun~Wu\textsuperscript{\rm 1}, 
    Josef~Kittler\textsuperscript{\rm 2},
    Tianyang~Xu\textsuperscript{\rm 1},
    Sara~Atito\textsuperscript{\rm 2},
    Muhammad~Awais\textsuperscript{\rm 2},  
    and Zhenhua~Feng\textsuperscript{\rm 2}
}
\affiliations{
    \textsuperscript{\rm 1}School of Artificial Intelligence and Computer Science, Jiangnan University, Wuxi 214122, China\\
    \textsuperscript{\rm 2}School of Computer Science and Electronic Engineering, and CVSSP, University of Surrey, Guildford GU2 7XH, UK   \\
    E-mail: congwu@stu.jiangnan.edu.cn, 
    wu\_xiaojun@jiangnan.edu.cn; tianyang\_xu@163.com;
    \{sara.atito, muhammad.awais, j.kittler, z.feng\}@surrey.ac.uk
}

\usepackage{bibentry}

\begin{document}

\maketitle

\begin{abstract}
Contrastive learning has achieved great success in skeleton-based action recognition. However, most existing approaches encode the skeleton sequences as entangled spatiotemporal representations and confine the contrasts to the same level of representation. Instead, this paper introduces a novel contrastive learning framework, namely Spatiotemporal Clues Disentanglement Network (SCD-Net). Specifically, we integrate the decoupling module with a feature extractor to derive explicit clues from spatial and temporal domains respectively. As for the training of SCD-Net, with a constructed global anchor, we encourage the interaction between the anchor and extracted clues. Further, we propose a new masking strategy with structural constraints to strengthen the contextual associations, leveraging the latest development from masked image modelling into the proposed SCD-Net. We conduct extensive evaluations on the NTU-RGB+D (60\&120) and PKU-MMD (I\&II) datasets, covering various downstream tasks such as action recognition, action retrieval, transfer learning, and semi-supervised learning. The experimental results demonstrate the effectiveness of our method, which outperforms the existing state-of-the-art (SOTA) approaches significantly.
\end{abstract}

\section{Introduction}
Skeleton-based action recognition focuses on identifying human actions via skeleton sequences, which has witnessed significant advancements in recent years. On one hand, deep networks, such as Graph Convolutional Network (GCN)~\cite{yan2018spatial}, have been investigated and successfully applied for the task at hand. On the other hand, several large-scale datasets, \textit{e.g.}, NTU-RGB+D~\cite{shahroudy2016ntu}, have been proposed, providing an experimental foundation for further development of the area. However, like most visual tasks, the training of a high-performance model typically requires a massive amount of high-quality labelled data. This requirement poses a significant challenge in data collection and annotation. Fortunately, self-supervised learning has emerged as a solution to address this challenge by leveraging inherent associations instead of relying on annotations. In particular, 
recent investigations~\cite{dong2023hierarchical} have demonstrated that contrastive learning, owing to its interpretability and transferability, has emerged as a front-runner in self-supervised skeleton-based action recognition.



However, several crucial aspects are disregarded by existing approaches.
First, the encoder is responsible for mapping the input into a latent space where the contrast can be conducted. 
While most previous methods~\cite{zhang2022contrastive,franco2023hyperbolic} concentrate on obtaining unified information through commonly used spatiotemporal modelling networks.
Their designs result in the complete entanglement of information, failing to provide clear indications for subsequent contrastive measures.
There have been sporadic attempts~\cite{dong2023hierarchical} aiming to extract absolutely isolated spatial or temporal information.
But repeated evidence has shown that complete isolation of spatiotemporal information is suboptimal for action recognition~\cite{kay2017kinetics,lin2019tsm}.
More importantly, most approaches focus on constructing contrast pairs at same level of representation~\cite{guo2022contrastive} during optimisation; Or attempt to force the interaction between information flows, overlooking the gap between domains~\cite{dong2023hierarchical}.
In addition, existing techniques~\cite{thoker2021skeleton} often limit themselves to scale transformation, which results in not fully capitalising on the potential of data augmentation.
Here we introduce a novel contrastive learning framework that focuses on disentangling spatiotemporal clues, and exploits masking in data augmentation to provide more discriminative inputs, thereby prompting the model to learn more robust interactions.



To leverage the intricate features present in skeleton sequences, we propose a dual-path decoupling encoder to generate explicit representations from spatial and temporal domains. 
Our encoder comprises two main subsystems: a feature extractor and a decoupling module.
The role of the feature extractor is to extract fundamental spatiotemporal features from skeleton sequences as the intermediate representations.
Since lacking an overall grasp of the skeleton sequence, it is difficult to obtain a picture of the features simply by modelling from a certain perspective.
Next, we generate token embeddings by projection and refine the sequence features with a transformer-based module.
The decoupling modules are instrumental to deriving disentangled joint-based and frame-based representations, leading to enhanced interpretability of the learned representations.

The principle underlying contrastive learning lies in achieving that the encoded query exhibits similarity with its corresponding key while showing dissimilarity with other keys from the backup queue~\cite{he2020momentum}.
Here we extend contrastive loss to measure discrimination among representations of multiple spatiotemporal granularities.
We strategically incorporate a global view of spatiotemporal representation as an anchor and evaluate its correlation with other representations obtained from an alternate encoder.
To elaborate further, we fuse and project the cues derived from the encoder into the contrastive space to create a global representation.
Our aim is to establish a bridge that facilitates the interaction of information across different domains through the utilisation of this anchor.

Furthermore, to prompt the model to learn more robust interactions, we propose an innovative mask-based data augmentation in a structurally-constrained manner. 
Specifically, we mask the adjacent area of the randomly selected region in the spatial domain and construct cube-based random masking in the temporal domain. 
This structured masking strategy serves to significantly increase the variety of training data. Moreover, it enables the model to implicitly capture spatiotemporal contextual relationships within the skeleton sequence.

\begin{figure}[t]
\begin{center}
\includegraphics[width=1.\linewidth]{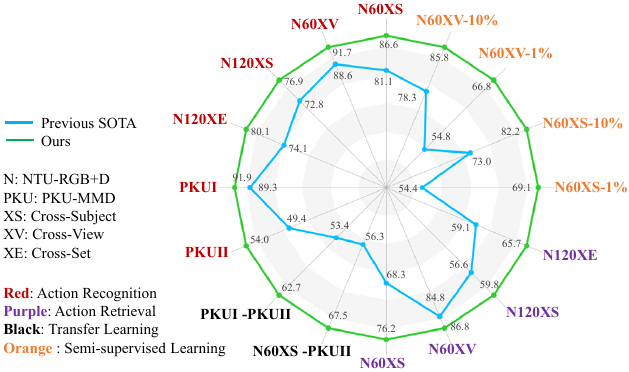}
\end{center}
\caption{A comparison of the proposed method with HiCo-Transformer~\cite{dong2023hierarchical}, using multiple evaluation metrics.
(Better view in colour.)
}
\label{fig: overall prefermance}
\vspace{-0.5em}
\end{figure}

We perform extensive experiments to demonstrate the effectiveness of the proposed method. 
As shown in Figure~\ref{fig: overall prefermance}, the results indicate that our approach surpasses the mainstream methods in all downstream tasks, demonstrating its superior capabilities in skeleton-based action understanding. 

In summary, the key innovations of SCD-Net include:
\begin{itemize}
\item A novel contrastive learning framework that is instrumental to decoupling spatiotemporal clues and enhancing the discriminatory capacity of action representations.
\item The novel decoupling encoders that are designed to extract clean spatial and temporal representations from complexly entangled information.
\item A new contrastive loss that encourages the network to capture meaningful interactions across different domains.
\item A structurally-constrained masking, which reflects the inherent properties of skeleton sequences in a conceptually unique way, is proposed for data augmentation.
\item SCD-Net redefines the new SOTA across all downstream tasks.
\end{itemize}


\section{Related Work}
\label{sec_rw}

\subsection{Skeleton-based Action Recognition}
Skeleton-based action recognition has garnered significant attention by the research community~\cite{ke2017new,gupta2021quo,duan2022pyskl}.
In earlier approaches~\cite{du2015hierarchical, chen2006human, zhu2019cuboid}, customised techniques were employed to classify skeletons via traditional feature extraction methods. 
Recently, GCN-based approaches~\cite{yan2018spatial, li2019actional, liu2020disentangling} have gained prominence in the field. 
The general paradigm initially models the skeleton sequence as a spatiotemporal graph and subsequently employs information aggregation and updating techniques. 
Inspired by the notable achievements of transformer~\cite{dosovitskiy2020image,liu2022video}, some recent methods~\cite{zhang2021stst,zhang2022zoom} have explored its powerful sequence modelling capability for skeleton-based tasks.

\subsection{Contrastive Learning}
Contrastive learning is a typical solution for self-supervised learning.
Unlike generative learning~\cite{zhu2020s3vae,huang2022self}, contrastive learning does not involve explicit generation or reconstruction of the input.
Instead, it focuses on learning discriminative representations through a contrastive loss.
Most contrastive learning methods ~\cite{chen2020simple,grill2020bootstrap} operate on the principle of pulling positive pairs closer to each other, while simultaneously pushing dissimilar pairs farther apart within a projection space. 
By exploring the internal properties within the data, contrastive learning enables learning  more generalised and robust representations, resulting in remarkable performance on downstream tasks~\cite{wang2022robustness}.

\subsection{Contrastive Learning for Skeleton-based Action Recognition}
Contrastive learning has also been successfully employed in skeleton-based action recognition.
Thoker \emph{et al.}~\cite{thoker2021skeleton} proposed the intra-skeleton and inter-skeleton contrastive loss, achieving promising results in several downstream tasks.
Dong \emph{et al.}~\cite{dong2023hierarchical} utilised down-sampling operations at different stages of the encoder to obtain multi-scale features for constructing a hierarchical contrastive learning framework.
Franco \emph{et al.}~\cite{franco2023hyperbolic} proposed a novel approach that involves projecting the encoded features into a hyperbolic space, which is a non-Euclidean space that allows more efficient modelling of complex association. 
Despite these advances, most existing studies overlook the crucial step of extracting and disentangling spatial and temporal clues from skeleton sequences,
not to mention the failure of considering the interactions among representations of different domains.

For contrastive learning, data augmentation processes the training sample to obtain positive input pairs with certain differences.
Thoker \emph{et al.}~\cite{thoker2021skeleton} used various spatiotemporal augmentation techniques, including pose augmentation, joint jittering, and temporal crop-resize, to generate different inputs for the query and key encoders.
While most methods follow similar scale transformation paradigms, 
Zhou \emph{et al.}~\cite{zhou2023self} proposed a strategy of masking selected nodes and frames, which greatly extends the augmentation to ''destroy'' the data structure.
However, unlike image data, skeleton sequences have strong physical associations, meaning that even if a certain node or frame is corrupted, it can easily be corrected using the information from adjacent areas~\cite{cheng2020decoupling}.
Incorporating the structural constraints, we expand the point-based masking approach to area-based masking. 
This extension aims to prevent potential data leakage and enhance the learning capabilities of SCD-Net.

\section{The Proposed SCD-Net}
\label{sec_method}
In this section, we will initially present the overall framework of SCD-Net, followed by a detailed introduction to each of its components in the subsequent sections.



\subsection{The Overall Framework}
\begin{figure}[!t]
\centering
\subfloat[Data augmentation.]{
\includegraphics[width=1.\linewidth]{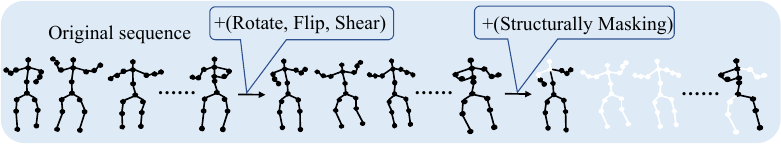}}

\subfloat[The framework of SCD-Net.]{
\includegraphics[width=1.\linewidth]{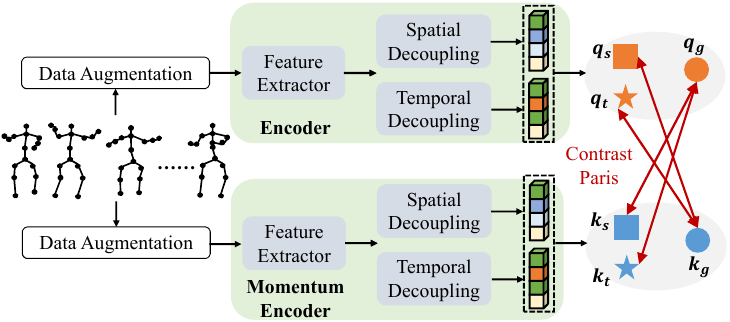}}
\caption{
Our model benefits from three innovations: a dual-path encoder for distinct spatiotemporal information decoupling; a bespoke cross-domains contrastive loss promoting the information interaction; a structurally-constrained masking strategy for efficient data augmentation.
}
\label{fig: overall framework}
\vspace{-0.5em}
\end{figure}

The overall pipeline of the proposed method consists of two branches, as shown in Figure~\ref{fig: overall framework} (b).
Each branch has the same components, including data augmentation and encoder. 
For any input data, we link the outputs obtained by the encoder and momentum encoder to form contrast pairs.






\begin{figure*}[t]
\centering
\includegraphics[width=1.\linewidth]{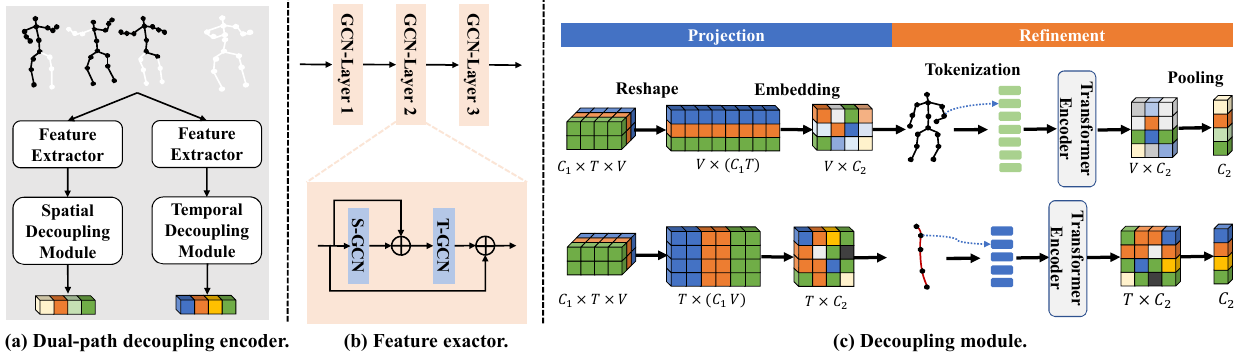}
\caption{The dual-path decoupling encoder that provides clean spatial and temporal representations of a skeleton sequence.
} 
\label{fig: encoder}
\vspace{-0.5em}
\end{figure*}

To elaborate further, the input of the network is defined as a sequence of human body key points, denoted as $\mathcal{X} \in \mathbb{R}^{C \times T \times V}$, where $T$ is the length of the sequence, $C$ is the physical coordinate defined in a 2D/3D space, $V$ is the number of key points. 
In SCD-Net, we first apply data augmentation to generate the augmented views for the  encoders. 
Second, 
for each encoder, we deploy feature extraction and (spatial/temporal) decoupling operations to generate  spatial feature $\mathbf{z}_s \in \mathbb{R}^{C_2}$ and temporal feature $\mathbf{z}_t \in \mathbb{R}^{C_2}$ from the entangled information.
Third, we project these clues into the same semantic space to obtain the final representations.

The loss function, $\mathcal {L}_{\mathbf{\theta},\mathbf{\xi}}$, is defined as a measure of interactions of these representations. 
The parameters $\theta$ and $\xi$ specify the architecture corresponding to the encoder and the momentum encoder.
During the optimisation,
the loss is back-propagated only through the encoder, while the parameters of the momentum encoder are updated using a momentum update strategy.
So the final optimiser is:
\begin{equation}
\boldsymbol{\theta}\leftarrow \operatorname{optimizer}(\boldsymbol{\theta}, \nabla _{\boldsymbol{\theta}}\mathcal {L}_{\boldsymbol{\theta},\boldsymbol{\xi}},r), 
\boldsymbol{\xi}=\boldsymbol{\xi}*m+\boldsymbol{\theta}*(1-m),
\label{eq1}
\end{equation}
where $r$ and $m$ are the learning rate and decay rate.

\subsection{The Dual-path Decoupling Encoder}
\label{secd}
In general, the features extracted from a skeleton sequence are characterised as complex spatiotemporal associations describing an action. 
However, we argue that this paradigm is not suitable for contrastive learning. 
As the information is greatly entangled, it is difficult to provide clear guidance for the subsequent comparison.
In SCD-Net, we advocate a dual-path decoupling encoder to extricate clear and multiple discriminative cues from the complex sequence information.
Such clues provide clear instructions for a subsequent contrast quantification.
More importantly, a reliable assessment of the contrast between different domains is likely to provide stronger discrimination.



For brevity, we generally denote the augmented input for the encoder as $\mathcal{X}$.
As demonstrated by the existing studies, completely isolating the information flow is sub-optimal~\cite{lin2019tsm,wang2021tdn}.
Given that, we apply a spatiotemporal modelling network to extract the intermediate features.
Inspired by the excellent performance in modelling skeleton sequences~\cite{yan2018spatial}, we use a $l_g$-layer GCN,
consisting of spatial-GCN (S-GCN) and temporal GCN (T-GCN), to obtain unified representations $\mathcal{Y}\in \mathbb{R}^{C_1 \times T\times V}$.
This can be expressed as a process of aggregation and updating of adjacent features.
Specifically, for any $\mathcal{X}_{ti}\in \mathbb{R}^{C}$, where $t$ and $i$ are the frame and joint index, the newly generated features $\mathcal{Y}_{ti}\in \mathbb{R}^{C_1}$  can be expressed as: 
\begin{equation}
\mathcal{Y}_{ti}=\sum_{\mathcal{X}_{uj}\in \operatorname{B}(\mathcal{X}_{ti})}\frac{1}{\operatorname{Z}_{ti}(\mathcal{X}_{uj})}\cdot \mathcal{X}_{uj}\cdot \operatorname{w}(\operatorname{l}_{ti}(\mathcal{X}_{uj})),
\label{eq2}
\end{equation}
where $\operatorname{B}(\mathcal{X}_{ti})$ denotes the kernel of the graph convolution operation on $\mathcal{X}_{ti}$, $\operatorname{Z}(\cdot)$ represents normalisation, $\operatorname{w}(\cdot)$ is the weight function, and $\operatorname{l}(\cdot)$ maps adjacent nodes to the corresponding subset index.

Given the intermediate spatiotemporal representation, $\mathcal{Y}$, the following step is decoupling operation, which involving projection and refinement,
as shown in Figure~\ref{fig: encoder}. 
Specifically, we
perform a dimension transformation on $\mathcal{Y}$ to derive $\mathcal{Y}_{rs}\in \mathbb{R}^{V\times (C_1T)}$ and $\mathcal{Y}_{rt}\in \mathbb{R}^{T\times (C_1V)}$. 
These transformed representations are then projected to higher semantic space to obtain the corresponding 
spatial and temporal
embeddings. 
For instance, the spatial embedding operation is defined as:
\begin{equation}
\mathcal{Y}_s=\mathcal{W}_{s2}*\operatorname{ReLU}(\mathcal{W}_{s1}*\mathcal{Y}_{rs}+\mathcal{B}_{s1})+\mathcal{B}_{s2}, \\
\label{eq3}
\end{equation}
where $\mathcal{W}$ and $\mathcal{B}$ are the trainable weights and bias, 
$\mathcal{Y}_{s}\in \mathbb{R}^{V\times C_2}$.
However, the current embedding is still a rough representation as the current features lack explicit interactions within points or frames.
While the feature extraction operation incorporates significant spatiotemporal interactions, these interactions often become intertwined. 
Hence, it remains crucial to address the interaction of individual spatial and temporal embeddings.
Here we use a $l_t$-layer self-attention network to construct the self-correlation information extraction process that refines the spatial and temporal representations,
as shown in Figure~\ref{fig: encoder}. 
The transformer architecture used in this method has two main components: self-attention and feed-forward modules. 
For instance, we obtain $\boldsymbol{z_s}\in \mathbb{R}^{C_2}$ as follows:
\begin{equation}
\mathcal{\hat{Z}}_{si}=\operatorname{SoftMax}\left[\frac{\left(\operatorname{F}_q(\mathcal{Y}_s)\right)_i\cdot \left(\operatorname{F}_k(\mathcal{Y}_s)\right)^{T}_i}{\sqrt{d_i}}\cdot \left(\operatorname{F}_v(\mathcal{Y}_s)\right)_i\right],
\nonumber
\end{equation}
\begin{equation}
\mathcal{\hat{Z}}_s=\operatorname{WM}[\mathcal{\hat{Z}}_{s1}, ..,\mathcal{\hat{Z}}_{si},..., \mathcal{\hat{Z}}_{sh}]+\mathcal{Y}_s,\quad i\in [0,h],
\nonumber
\end{equation}
\begin{equation}
\mathbf{z}_s = \operatorname{MaxPooling}[\operatorname{F}_c^u(\operatorname{FFN}(\operatorname{LN}(\mathcal{\hat{Z}}_s))+\mathcal{\hat{Z}}_s],
\label{eq4}
\end{equation}
where $\operatorname{F}$ represents feature projection, implemented by a fully connected layer, with $[\dots]$ denoting the concatenation operation, and $h$ signifying the number of heads.
$\boldsymbol{z_t}\in \mathbb{R}^{C_2}$ can also be obtained by similar operations.

\subsection{Cross-domain Contrastive Loss}
With the decoupled spatial and temporal representations, 
as shown in Figure~\ref{fig: overall framework}, we first obtain the final representations by:
\begin{equation}
\mathbf{q}_s=\operatorname{F}_s(\mathbf{z}_s), \quad
\mathbf{q}_t=\operatorname{F}_t(\mathbf{z}_t).
\label{eq5}
\end{equation}
where $\operatorname{F}_s$, $\operatorname{F}_t$  are the corresponding projection functions, which can be defined by two fully connected layers, similar to Eq.~(\ref{eq3}).
As we discussed earlier, there is an obvious gap between spatial and temporal domains, for which we introduce a global perspective $\mathbf{q}_g$ compatible with both as a intermediary for contrasts.
\begin{equation}
\mathbf{q}_g=\operatorname{F}_g[\mathbf{z}_t, \mathbf{z}_s],
\label{eq5}
\end{equation}
where $\operatorname{F}_g$  are the corresponding projection function.
The outputs ($\mathbf{k}_s$, $\mathbf{k}_t$, $\mathbf{k}_g$) of the corresponding key encoder, can also be obtained by a similar process.

Based on these candidate features, we define a new cross-domain loss. 
The core of our design lies in anchoring the global representation and building its association with other representations obtained by another encoder.
The loss function is defined as,
\begin{equation}
\begin{split}
\mathcal{L}_{\mathbf{\theta},\mathbf{\xi}}
\triangleq \lambda _1\cdot \mathcal L(\mathbf{q}_g, \mathbf{k}_s)+ \lambda _2\cdot \mathcal L(\mathbf{q}_g, \mathbf{k}_t)+\\ \lambda _3\cdot \mathcal L(\mathbf{q}_s, \mathbf{k}_g) +\lambda _4\cdot \mathcal L(\mathbf{q}_t, \mathbf{k}_g),
\end{split}
\label{eq6}
\end{equation}
where $\lambda$ is the mixing weight of the averaging operation. 
Specifically, 
for any given contrast pair $\mathbf{u}$ and $\mathbf{v}$,
$\mathcal{L}(\mathbf{u}, \mathbf{v})$ evaluates the correlation between $\mathbf{u}$ and $\mathbf{v}$.
The objective is to minimise the distance between positive pairs from the query and key encoders, while maximising the distance from the other features.

To achieve this, we employ the contrastive loss based on InfoNCE~\cite{oord2018representation} as follows:
\begin{equation}
\mathcal L(\mathbf{u}, \mathbf{v})=
-\log\frac{\operatorname{h}(\mathbf{u}, \mathbf{v})}{\operatorname{h}(\mathbf{u}, \mathbf{v})+\sum_{\mathbf{m}\in {M}}{\operatorname{h}(\mathbf{u},\mathbf{m})}},
\label{eq7}
\end{equation}
where $\operatorname{h}(\mathbf{u},\mathbf{v})=\operatorname{exp}(\mathbf{u}\cdot \mathbf{v}/\tau)$ is the exponential similarity measurement.
We denote the first-in-first-out queue of the previously extracted features, containing $l_m$ negative samples, by ${M}$. 

\begin{table*}[!th]
\caption{A comparison of the proposed method with the mainstream methods in action recognition. 
The {\color{blue}{blue}} font indicates the previous SOTA.
}
\begin{center}
\begin{threeparttable}
\resizebox{.8\width}{!}{
\begin{tabular}{c c c c c c c c}
\hline
\multirow{2}{*}{\textbf{Method}}  & \multirow{2}{*}{\textbf{Encoder}}   & \multicolumn{2}{c}{\textbf{NTU-60}} & \multicolumn{2}{c}{\textbf{NTU-120}} &PKU-MMD I &PKU-MMD II \\ 
   &          &x-sub         & x-view     &x-sub         &x-setup  &x-sub        &x-sub \\ \hline
\multicolumn{1}{c}{\textbf{\emph{Encoder-decoder}}} \\
LongT GAN~\cite{zheng2018unsupervised}(\scriptsize  AAAI’18)  & GRU &52.1 &56.4 &-&- &67.7 &26.5 \\
EnGAN-PoseRNN~\cite{kundu2019unsupervised}(\scriptsize WACV’19)  &LSTM &68.6 &77.8&-&- &-&-\\
H-Transformer~\cite{9428459}(\scriptsize  ICME’21)  &Transformer &69.3 &72.8&-&-&-&-\\
SeBiReNet~\cite{nie2020unsupervised}(\scriptsize  ECCV’20) &GRU &- &79.7&-&-&-&-\\
Colorization~\cite{yang2021skeleton}(\scriptsize  ICCV’21)  &GCN &75.2 &83.1&-&-&-&-\\
GL-Transformer~\cite{kim2022global}(\scriptsize  ECCV’22)  &Transformer &76.3 &83.8&66.0&68.7&-&-\\ \hline
\multicolumn{1}{c}{\textbf{\emph{Hybrid learning}}} \\
MS$^2$L~\cite{lin2020ms2l}(\scriptsize  ACMMM’21)  &GRU&52.6 &-&-
&- &64.9 &27.6\\
PCRP~\cite{xu2021prototypical}(\scriptsize TMM’21) &GRU&54.9 &63.4&43.0&44.6&-&-\\ \hline
\multicolumn{1}{c}{\textbf{\emph{Contrastive-learning}}} \\
CrosSCLR~\cite{Li_2021_CVPR}(\scriptsize  CVPR’21)  &GCN &72.9 &79.9&-&- &84.9\tnote{*} &27.6\tnote{*}\\
AimCLR~\cite{guo2022contrastive}(\scriptsize  AAAI’22)  &GCN &74.3 &79.7&63.4&63.4&87.8\tnote{*} & 38.5\tnote{*}\\
ISC~\cite{thoker2021skeleton}(\scriptsize  ACMMM’21) &GRU\&CNN\&GCN &76.3 &85.2&67.1&67.9& 80.9&36.0\\
HYSP~\cite{franco2023hyperbolic}(\scriptsize  ICLR’23)  &GCN &78.2 &82.6&61.8&64.6&83.8&-\\
SkeAttnCLR~\cite{hua2023part}(\scriptsize  IJCAI’23)  &GCN &80.3 &86.1&66.3&{\color{blue}{74.5}}&87.3&{\color{blue}{52.9}}\\
ActCLR~\cite{lin2023actionlet}(\scriptsize  CVPR’23)  &GCN &80.9 &86.7&69.0&70.5&-&-\\
HiCo-Transformer~\cite{dong2023hierarchical}(\scriptsize AAAI’23)  &Transformer & {\color{blue}{81.1}} &{\color{blue}{88.6}}&{\color{blue}{72.8}}&74.1 &{\color{blue}{89.3}}&49.4\\\hline
SCD-Net (Ours) &GCN\&Transformer &\textbf{86.6} &
\textbf{91.7} &\textbf{76.9} &\textbf{80.1} &\textbf{91.9} &\textbf{54.0} \\
\hline 
\end{tabular}}
\end{threeparttable}
\label{tab: actionrecognition}
\end{center}
\vspace{-0.5em}
\end{table*}

\subsection{Data Augmentation}
\label{sec3.3}
By imposing structural constraints, our approach applies the masking operation within a local region around the current randomly selected joints or frames instead of relying only on isolated points or frames. 
In this way, we substantially eliminate explicit local contextual associations,
and force the encoders to model robust contextual relationships through interactive contrastive learning.

\subsubsection{Structurally Guided Spatial Masking}
Considering the physical structure of skeleton, when a certain joint is selected for masking, we simultaneously mask the points in its adjacent area.
Let us represent the adjacency relationship using the matrix $\mathbf{P}$. $\mathbf{P}_{ij}=1$, if joints $i$ and $j$ are connected, otherwise $\mathbf{P}_{ij}=0$.
We denote $\mathbf{D}=\mathbf{P}^n$, where $n$ is the exponent.
The element $\mathbf{D}_{ij}$ in $\mathbf{D}$ represents the number of paths that can be taken to reach node $j$ from node $i$ by walking $n$ steps. Note that reversal and looping are allowed.
To impose a structural constraint, when node $i$ is selected, we perform the same augmentation operation on all nodes $j$ for which $\mathbf{D}_{ij}\neq 0$.
The only undesirable artefact of this operation is that it may give rise to a  variable number of candidate joints.
To avoid this, for several randomly selected nodes, the actual augmentation is applied only to a fixed number ($k$) of points exhibiting the highest overall response on $\mathbf{D}$.

\subsubsection{Cube-based Temporal Masking}
The sequence follows a linear relationship in time.
To avoid information leakage between adjacent frames~\cite{tong2022videomae}, we construct a cube, defined by a selected segment and its adjacent frames. 
Specifically, we start by dividing the input sequence into $s$ cubes of equal length. Next, we randomly select $r$ cubes as the candidates for masking.

We denote the data augmentation candidates as $\mathcal T$. 
Given a skeleton sequence $\mathcal{X}$, the augmented view is obtained by: 
\begin{equation}
\mathcal{X}^a \triangleq \operatorname{t_n}(\mathcal{X}^a_{n},p_n)\dots \triangleq \operatorname{t_1}(\mathcal{X}^a_1,p_1),
\label{eq10}
\end{equation}
where $\mathcal{X}^a_1=\mathcal{X}$, $t_1, \dots, t_n\sim \mathcal T$, and 
if $p=\operatorname{False}$, $\operatorname{t}$ degenerates into an identity map.

\section{Experiments}
\label{sec_exp}

\subsection{Experimental Settings}
\subsubsection{Datasets} 
We evaluate the proposed method on four bench-marking datasets, NTU-RGB+D (60\&120)~\cite{shahroudy2016ntu} and PKU-MMD (I\&II)~\cite{liu2017pku}.

\subsubsection{Implementation Details}
For the input data, 64 frames are randomly selected for training and evaluation.
We perform data augmentation operations, including rotate, flip and shear, as well as the proposed structural spatial masking and temporal masking, on the selected sequence. 
Each operation has a $50\%$ chance of being executed.
For masking, we set $n=2$, $k=8$, $s=16$, $r=6$.
For the encoder, we refer to MoCo~\cite{he2020momentum} and build a query encoder and the corresponding key encoder. 
The two encoders have exactly the same structure as shown in Figure~\ref{fig: encoder}. 
For feature extractor, we borrow the structure from CTR-GCN~\cite{chen2021channel} as the basic operation.
For network optimisation, we set the queue length of $M$ to 8192 (except 2048 for PKU-MMD I), moco momentum to 0.999, and softmax temperature to 0.2. 

More details of the datasets and implementation are presented in the supplementary materials. 

\subsection{Comparison with The SOTA Methods}
To evaluate the merits of the proposed SCD-Net, we construct multiple downstream tasks, including action recognition, action retrieval, transfer learning and semi-supervised learning.

\paragraph{Action Recognition}
Here we adopt the linear evaluation method, which involves fixing the pre-trained parameters and training only a fully connected layer for label prediction. Table~\ref{tab: actionrecognition} presents a comparison of our approach with other SOTA methods on several popular datasets. 
The results demonstrate that our method outperforms all the existing approaches by a large margin. 
Specifically, we achieve 5.5\% and 3.1\% improvements over the previous best method on NTU-60 x-sub and x-view, respectively. 
On NTU-120, our approach surpasses the previous SOTA by 4.1\% and 5.6\% on x-sub and x-set, respectively.
Again, for the PKU-MMD dataset, SCD-Net achieves 91.9\% on V1 and 54.0\% on V2, which are much higher than the existing SOTA results.

\begin{table}[!t]
\centering
\caption{A comparison with the mainstream methods in action retrieval.
The {\color{blue}{blue}} font indicates the previous SOTA.
}
\tabcolsep=1.5mm
\resizebox{.8\width}{!}{
\begin{tabular}{c c c c c}
\hline
\multirow{2}{*}{\textbf{Method}}   & \multicolumn{2}{c}{\textbf{NTU-60}} & \multicolumn{2}{c}{\textbf{NTU-120}} \\ 
    &x-sub         & x-view     &x-sub         &x-setup  \\ \hline
LongT GAN~\cite{zheng2018unsupervised} &52.1 &56.4 &-&-\\
P\&C~\cite{su2020predict} &50.7 &76.3 &39.5&41.8\\
AimCLR~\cite{guo2022contrastive} & 62.0&-&-&-\\
ISC~\cite{thoker2021skeleton}  &62.5 &82.6&50.6&52.3\\
SkeAttnCLR~\cite{hua2023part} &{\color{blue}{69.4}} &76.8&46.7&58.0\\
HiCo-Transformer~\cite{dong2023hierarchical}  &68.3 &{\color{blue}{84.8}}&{\color{blue}{56.6}}&{\color{blue}{59.1}}\\
\hline
SCD-Net (Ours) &\textbf{76.2} &\textbf{86.8} &\textbf{59.8} &\textbf{65.7} \\
\hline 
\label{tab: retrei}
\end{tabular}}
\vspace{-0.5em}
\end{table}

\begin{table}[!t]
\centering
\caption{A comparison with the mainstream methods in transfer learning.
The {\color{blue}{blue}} font indicates the previous SOTA.
}
\resizebox{.8\width}{!}{
\begin{tabular}{c c c}
\hline
\multirow{2}{*}{\textbf{Method}} & \multicolumn{2}{c}{\textbf{Transfer to PKU-MMD II}} \\ 
&PKU-MMD I    &NTU-60  \\ \hline
LongT GAN~\cite{zheng2018unsupervised} &43.6 &44.8\\
MS$^2$L~\cite{lin2020ms2l} &44.1 &45.8\\
ISC~\cite{thoker2021skeleton} & 45.1&45.9\\
HiCo-Transformer~\cite{dong2023hierarchical} &{\color{blue}{53.4}}&{\color{blue}{56.3}}\\
\hline
SCD-Net (Ours)  &\textbf{62.7} &\textbf{67.5} \\
\hline 
\label{tab: trans}
\end{tabular}}
\vspace{-0.5em}
\end{table}

\begin{table}[ht]
\centering
\caption{A comparison with the mainstream methods in semi-supervised learning.
The {\color{blue}{blue}} font indicates the previous SOTA.
}
\resizebox{.8\width}{!}{
\begin{tabular}{c c c c c}
\hline
\multirow{2}{*}{\textbf{Method}}   & \multicolumn{2}{c}{\textbf{x-sub}} & \multicolumn{2}{c}{\textbf{x-view}} \\ 
    &1\%         & 10\%      &1\%         & 10\%  \\ \hline
LongT GAN~\cite{zheng2018unsupervised} &35.2 &62.0 &-&-\\
MS$^2$L~\cite{lin2020ms2l} &33.1 &65.2 &-&-\\
ASSL~\cite{si2020adversarial} & -&64.3&-&69.8\\
ISC~\cite{thoker2021skeleton}  &35.7 &65.9&38.1&72.5\\
MCC~\cite{su2021modeling}  &- &60.8&-&65.8\\
Colorization~\cite{yang2021skeleton} &48.3 &71.7&52.5&78.9\\
CrosSCLR~\cite{Li_2021_CVPR} &- &67.6&-&73.5\\
HI-TRS~\cite{Li_2021_CVPR} &- &70.7&-&74.8\\
GL-Transformer~\cite{kim2022global} &- &68.6&-&74.9\\
HiCo-Transformer~\cite{dong2023hierarchical}  &{\color{blue}{54.4}} &{\color{blue}{73.0}}&{\color{blue}{54.8}}&{\color{blue}{78.3}}\\
\hline
SCD-Net (Ours) &\textbf{69.1} &\textbf{82.2} & \textbf{66.8}&\textbf{85.8} \\
\hline 
\end{tabular}}
\label{tab:semi}
\end{table}

\paragraph{Action Retrieval}
Referring to~\cite{thoker2021skeleton}, we use the KNeighbors classifier~\cite{cover1967nearest} for action retrieval while keeping all the pre-trained parameters fixed.
As reported in Table~\ref{tab: retrei}, our SCD-Net achieves promising results on the NTU-60's x-sub and x-view datasets, with the accuracy of 76.2\% and 86.8\%, respectively. Additionally, on the NTU-120's x-sub and x-set datasets, our method attains the accuracy of 59.8\% and 65.7\%, surpassing all the existing methods by a significant margin.

\paragraph{Transfer Learning}
For transfer learning, 
follow~\cite{dong2023hierarchical}, we apply the knowledge representation learned from one domain to another domain. Specifically, we load the pre-trained parameters from the PKU-MMD I and NTU-60 datasets respectively,
and fine-tune the model on the PKU-MMD II dataset, following the cross-subject evaluation protocol. 
The results presented in Table~\ref{tab: trans} demonstrate that our SCD-Net brings a performance improvement of 9.3\% and 11.2\%, as compared with the current SOTA results.

\paragraph{Semi-supervised Learning}
For semi-supervised learning, we first load the pre-trained parameters and then fine-tune the entire network on a partially labelled training set. 
In our experiment, we randomly select limited of labelled samples from the NTU-60 dataset for further training. 
The results in Table~\ref{tab:semi} show that even when only 1\% of the labels are available, our method achieves the accuracy of 69.1\% and 66.8\% on x-sub and x-view, respectively.
With 10\% of the labelled data available, the performance of our model is further improved
to 82.2\% and 85.8\%.


\subsection{Ablation Study}
In this part, we verify all the innovative components of the proposed SCD-Net.
All the experimental results are focused on the action recognition task using cross-subject evaluation on the NTU-60 dataset.

\paragraph{The Decoupling Encoder}
The primary role of our novel encoder is to extract crucial spatial and temporal representations. 
In Table~\ref{tab: encoder}, 
when we discard the feature extractor, the performance drops a lot.
This shows that the way of extracting completely isolated information flow is not feasible in the current task, which is also in line with our expectation.
It is worth noting that using non-shared feature extractors for the two branches leads to better performance than using a shared one. 
When we attempt to discard decouple module, compared with the default setting, the accuracy is decreased from 86.6\% to 63.7\% as the output is impacted by the spatiotemporal entanglement.
This situation improves after 
converting spatiotemporal representations into temporal and spatial domain-specific embeddings, resulting in an accuracy of 84.0\%. 
However, it was still inferior to the design with the refinement model.
This is because refinement provides powerful sequence modelling capabilities, thereby refining the current rough representations.

\begin{table}[t]
\centering
\caption{Ablation experiments with the decoupling encoder.
}
\resizebox{.8\width}{!}{
\begin{tabular}{c c c c}
\hline
\multirow{2}{*}{\textbf{Feature Extract}}&\multirow{2}{*}{\textbf{Decoupling}} &\multicolumn{2}{c}{\textbf{Accuracy}} \\
 & &Top-1 &Top-5 \\ \hline
Non-Shared & Projection \& Refinement &\textbf{86.6} &\textbf{97.6}   \\ 
None & Projection \& Refinement &80.0  &94.9   \\
Shared & Projection \& Refinement &85.2  &96.0   \\ \hline
Non-Shared &None & 63.7 &89.6   \\
Non-Shared & Projection &{84.0} &{97.1}   \\ 
\hline
\end{tabular}
}
\label{tab: encoder}
\vspace{-0.5em}
\end{table}

\begin{table}[t]
\centering
\caption{
The details of the encoder.
 The bold part in black indicates the best performance.}
\resizebox{.8\width}{!}{
\begin{tabular}{c c c c c c c}
\hline
\multicolumn{2}{c}{\textbf{GCN}}  & \multicolumn{3}{c}{\textbf{Trans}}  & \multicolumn{2}{c}{\textbf{Accuracy}}\\ 
Layer & Channel & Layer & Head & Channel &Top-1 & Top-5 \\ \hline
{2} & 64 & 1 & 8 & 2048 &85.4 &97.2 \\
{\textbf{3}} & \textbf{64} & \textbf{1} & \textbf{8} & \textbf{2048} &\textbf{86.6} & \textbf{97.6} \\
{4} & 64 & 1 & 8 & 2048 &86.6 &97.5 \\ 
3 & {32} & 1 & 8 & 2048 &86.4 &97.4 \\
3 & {128} & 1 & 8 & 2048 &85.8 &97.5 \\ \hline
3 & 64 & {2} & 8 & 2048 &86.4 &97.7 \\ 
3 & 64 & 1 & {4} & 2048 &86.2 &97.4 \\ 
3 & 64 & 1 & {16} & 2048 &85.7  &97.3\\
3 & 64 & 1 & 8 & {1024} &85.7 &97.5\\
3 & 64 & 1 & 8 & {4096} &85.1 &97.2\\ 
\hline 
\end{tabular}}
\vspace{-0.5em}
\label{tab: para}
\end{table}

\paragraph{Encoder Parameters}
In Table~\ref{tab: para}, we investigate the impact of the parameter settings on the model performance. 
Overall, the optimal performance is achieved when we use a 3-layer GCN block, 64 as the number of output channels, and set the transformer with 1 layer, 8 heads, and 2048 output channels.
The results also demonstrate that changing the parameters does not significantly affect the model's performance, indicating the stability of our approach. 
Additionally, we can see that the network size does not necessarily improve the performance, suggesting that it is not dependent on the network size.

\paragraph{Loss Function}
We report the results of different configurations of the loss function in Table~\ref{tab: loss}.
We can see that the interactive loss performs better than the traditional instance loss, leading to 0.7\% and 1.6\% performance boost.
When using all three granularities jointly, the model achieves optimal performance. 
This is because there is a significant gap between the nature of the video information conveyed by the spatial and temporal features, although they  describe the same action. 
The spatiotemporal information provides more comprehensive representations, which bridge this gap and enhances the discriminative ability. 
It is worth noting that the use of both loss functions jointly does not improve the performance further. 
This could be attributed to the fact that the supervisory information across the information flow already provides adequate guidance, and further guidance mechanisms are unnecessary. 

\begin{table}[!t]
\centering
\caption{A comparison of different loss functions.
'S', 'T', 'G' represent Spatial, Temporal and Global representations.
}
\resizebox{.8\width}{!}{
\begin{tabular}{ c c c c}
\hline
\multirow{2}{*}{\textbf{Granularity}} & \multirow{2}{*}{\textbf{Loss Type}} & \multicolumn{2}{c}{\textbf{Accuracy}} \\ 
& &Top-1 &Top-5\\ \hline
S-T & Instance Loss &84.6 &97.1\\
S-T & Interactive Loss &85.3 & 97.2 \\
S-T-G & Instance Loss &85.0 &97.5\\
S-T-G & Interactive Loss  &\textbf{86.6} & \textbf{97.6}  \\
S-T-G & Interactive Loss \& Instance Loss &86.5 &97.6 \\
\hline 
\end{tabular}}
\label{tab: loss}
\end{table}

\begin{table}[!t]
\centering
\caption{A comparison of different data augmentation strategies.}
\resizebox{.8\width}{!}{
\begin{tabular}{c c c}
\hline
\multirow{2}{*}{\textbf{Data Augmentation Strategy}}  & \multicolumn{2}{c}{\textbf{Accuracy}} \\
    &Top-1 &Top-5 \\ \hline
None &70.1&91.1 \\
$\operatorname{Conventional}$ &85.4&97.4 \\
$\operatorname{Conventional}, \operatorname{Spatial\ Mask}, \operatorname{Temporal\ Mask}$ &\textbf{86.6} & \textbf{97.6}  \\
$\operatorname{Conventional}, \operatorname{Random\ SpatioTemporal\ Mask}$ &85.6 &97.4   \\
$\operatorname{Conventional}, \operatorname{Spatial\ Mask}$ &86.2 &97.5  \\
$\operatorname{Conventional}, \operatorname{Temporal\ Mask}$ &83.7 &96.7   \\
$\operatorname{Spatial\ Mask}, \operatorname{Temporal\ Mask}$ &76.0 & 93.6  \\
\hline
\end{tabular}}
\label{tab: aug}
\vspace{-0.5em}
\end{table}



\paragraph{Data Augmentation}
Here we investigate the impact of different data augmentation strategies on the model performance. 
The results are reported in Table~\ref{tab: aug}.
Without any augmentation, the performance drops by more than 16\%, compared to the default setting.
When using only the conventional augmentation methods, including rotation, flipping, and shearing, the model achieves an accuracy of 85.4\%. 
After introducing the proposed structurally guided spatiotemporal augmentation, the performance of the model increases 1.2\% further.
Even with random masking, the performance is still lower than the default setting.

It is worth noting that discarding either spatial or temporal masking leads to a performance degradation.
Also, when only masking is used, the performance of the model is mediocre, even far worse than using only the conventional data augmentation methods.
That is because  our method performs a compensation, instead of replacement.
A proper masking further improves the diversity of the input data and promotes the model in learning more robust spatiotemporal context associations.
When combining all these techniques, the performance of the model is the best.

\begin{figure}[!t]
\centering
\subfloat[]{
\includegraphics[scale=0.2]{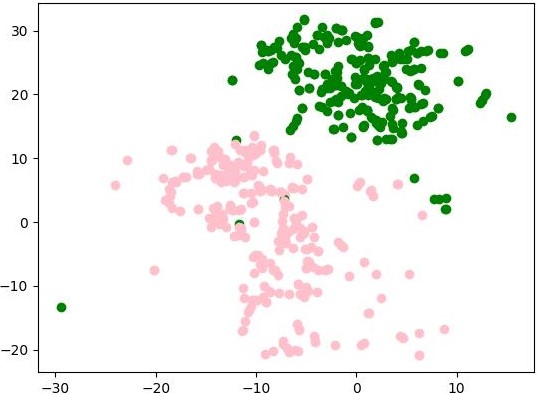}}
\subfloat[]{
\includegraphics[scale=0.2]{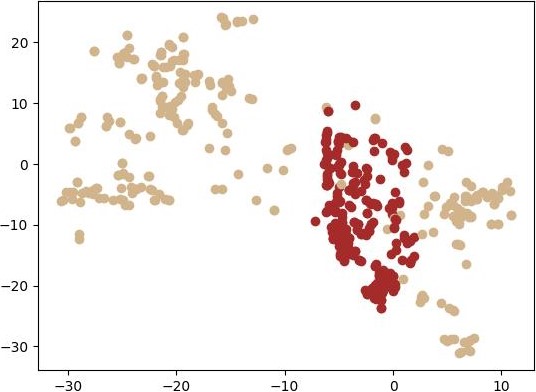}}
\subfloat[]{
\includegraphics[scale=0.2]{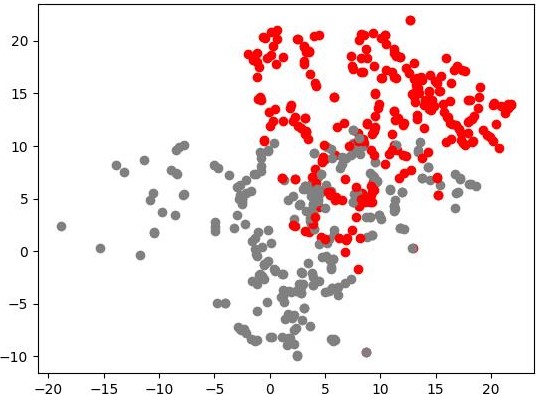}}

\subfloat[]{
\includegraphics[scale=0.2]{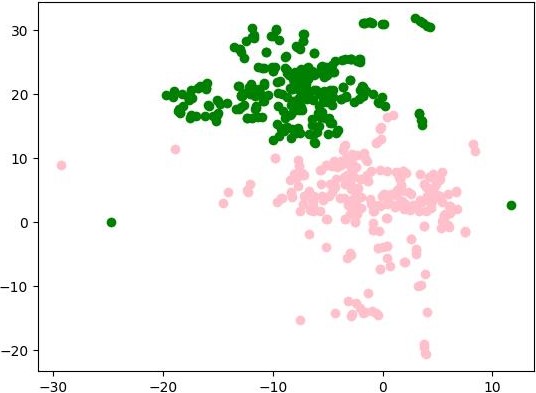}}
\subfloat[]{
\includegraphics[scale=0.2]{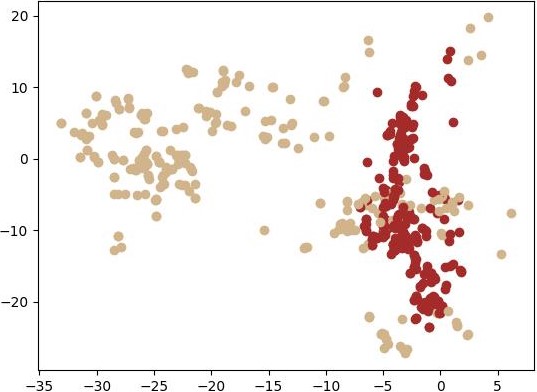}}
\subfloat[]{
\includegraphics[scale=0.2]{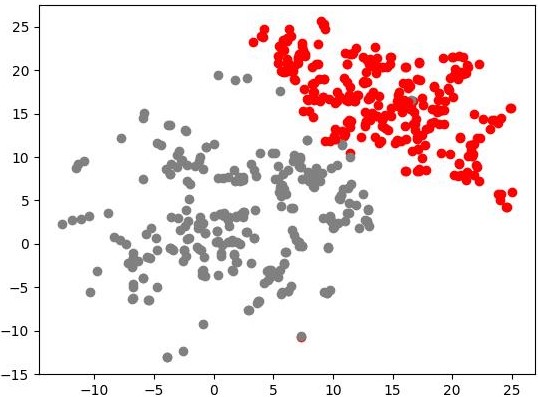}}
\caption{
Visualisation of decoupled clues.
Spatial clue: (a) 'throw' vs 'clapping'; (b) 'brush teeth' vs 'brush hair'; (c) 'drop' vs 'pick up'.
Temporal clue: (d) 'throw' vs 'clapping'; (e) 'brush teeth' vs 'brush hair'; (f) 'drop' vs 'pick up'; 
}
\label{fig: clues}
\vspace{-0.5em}
\end{figure}


\paragraph{Visualisation of The Decoupled Clues}
As shown in Figure~\ref{fig: clues}, we use t-SNE~\cite{van2008visualizing} to analyse the decoupled clues from SCD-Net. 
We select three groups of data with different emphases for comparison.
The first row represents the spatial clue and the second one is the temporal clue.
We can notice that from (a) and (d), 'throw’ and 'clapping’ have great separability on spatially and temporally.
From (b) and (e), ’brush teeth’ vs ’brush hair’ are more separable on spatial domain, because the most significant difference is the object.
According to (c) and (f), 'drop’ and 'pick up’ are more separable on temporal domain, while showing certain entanglement on the spatial domain, as they are in reverse order in temporally.
More importantly, the results demonstrate that our encoder successfully decouples the corresponding features, which makes the specificity between different cues correspond to the same samples.


\section{Conclusion}
\label{sec_con}
In this paper, we presented a new contrastive learning framework for unsupervised skeleton-based action recognition.
The key innovation is the design of spatiotemporal clue extraction mechanism.
In the proposed method, we first used a spatiotemporal modelling network to encode an action sequence, followed by a decoupling module for obtaining pure spatial and temporal representations.
An cross-domain loss was proposed to guide the learning of discriminative representations conveyed by different representations. 
The training of the system was facilitated by a novel data augmentation method tailored for the proposed unsupervised learning framework.
This method imposes structural constraints on action data perturbations to enhance the efficacy of contextual modelling and increase the diversity of the data.
Extensive experimental results obtained on widely used benchmarking datasets demonstrated the merits of the proposed method that defines a new SOTA of the area.

\section*{Acknowledgements}
This work was supported in part by the National Natural Science Foundation of China (Grant No. U1836218, 62020106012, 62106089, 61672265), the 111 Project of Ministry of Education of China (Grant No.B12018), the Postgraduate Research \& Practice Innovation Program of Jiangsu Province (Grant No. KYCX22\_2299), and the EPSRC Grants (EP/V002856/1, EP/T022205/1).

\bibliography{aaai24}
\clearpage
\section{Supplementary Material}
In this section, we delve into further details about our SCD-Net framework and offer additional details about the experimental settings. Moreover, we present a more extensive range of experimental results that highlight the efficacy of our novel contributions.

\section{More Details about SCD-Net Framework}
\paragraph{The architecture of decoupling encoder}
The decoupling encoder comprises feature extractor and spatial/temporal decoupling module. The final representation is subsequently acquired through a pooling operation. The specific network architecture and dimension transformation within this process are detailed in Table~\ref{table:encoder}.

\begin{table}[h]
\caption{The implementation details of our proposed decoupling encoder. The dimension of the output features is defined by the number of channels, frames, and joints, respectively.}
\centering
\resizebox{.8\width}{!}{
\begin{tabular}{ccc}
\hline 
\multicolumn{2}{c}{\textbf{Stage}} & \textbf{Output} \\
\hline
\multicolumn{2}{c}{Input} 		&{$3\times64\times25$} 	\\\hline
\multirow{3}{*}{Feature extractor}  &Layer1 & $64\times64\times25$\\
                        &Layer2 & $256\times64\times25$\\
                        &Layer3 & $64\times64\times25$\\\hline
\multirow{3}{*}{Spatial decoupling} &{Reshape}		&$1600\times64$	 \\
 &{Embedding}		&$2048\times64$	 \\
 &{Transformer} 	&$2048\times64$	 \\\hline
 \multirow{3}{*}{Temporal decoupling} &{Reshape}		&$4095\times25$	 \\
 &{Embedding}		&$2048\times25$	 \\
 &{Transformer} &$2048\times25$	 \\\hline
\multicolumn{2}{c}{Pooling} &$2048$	 \\\hline
\end{tabular}}
\label{table:encoder}
\end{table}



\paragraph{Data Augmentation}
As outlined in Listing 1, we provide the implementation particulars of our masking strategy.
The initial step of our masking approach involves the selection of the target region to be masked.
In the case of spatial masking, we start by choosing 5 points from the total of 25 joint points. Following this, using the adjacency matrix, we identify all nodes within a 2-unit distance from these 5 points. Subsequently, we opt for 8 nodes from this candidate set, based on their distances to original 5 points.
For temporal masking, we initially partition the existing 64 frames evenly into 16 clips. From these, we randomly choose frames corresponding to 4 of these clips.
After the identification of the masking region, the subsequent step involves masking these chosen points or frames to a value of 0.

\begin{lstlisting}[language=Python, caption=Python Code for Our Masking]
import numpy as np
def spatial_masking(input_data, A):
    # input_data: C T V M
    # A: Adjacency Matrix
    A = np.matmul(A, A)
    shuffle_index = np.random.randint(low=0, high=25, size=5)
    flag = A[shuffle_index].sum(0)
    joint_indicies = flag.argsort()[-8:]
    out = input_data.copy()
    out[:, :, joint_indicies, :] = 0
    return out
def temporal_masking(input_data):
    input_data = rearrange(input_data, 'c (t d) v m -> c t d v m', d=4)
    temporal_indicies = np.random.choice(16, 6, replace=False)
    out = input_data.copy()
    out[:, temporal_indicies] = 0
    out = rearrange(out, 'c t d v m -> c (t d) v m')
    return out
\end{lstlisting}

\begin{table*}[t]
\centering
\caption{Training details for the Self-supervised Learning and Action Recognition.}
\resizebox{.8\width}{!}{
\begin{tabular}{c|c c c c}
\hline
\multirow{2}{*}{Dataset and Tasks} & \multicolumn{2}{c}{Self-supervised Learning} & \multicolumn{2}{c}{Action Recognition}\\\cline{2-5}
&NTU\&PKU-MMD I  & PKU-MMD II &NTU\&PKU-MMD I  & PKU-MMD II   \\ \hline
Optimizer & \multicolumn{4}{c}{SGD}\\
Optimizer momentum& \multicolumn{4}{c}{0.9}\\\hline
Contrast queue &8192 &2048 &- &-\\
Contrast momentum &\multicolumn{2}{c}{0.999}&- &-\\
Contrast softmax temperature &\multicolumn{2}{c}{0.2}&- &-\\
Contrast dimension &\multicolumn{2}{c}{128}&- &-\\\hline
Weight decay& 0.0001 & 0.001 & 0 & 0.001\\
Initial learning rate &\multicolumn{2}{c}{0.01} &2 &0.002\\
Learning rate schedule &\multicolumn{2}{c}{[350]} &\multicolumn{2}{c}{[50,70]}\\
Batch size &\multicolumn{2}{c}{64} &1024 &16\\
Training steps & \multicolumn{2}{c}{450} &\multicolumn{2}{c}{80}\\ \hline
\end{tabular}}
\label{table: training}
\end{table*}

\begin{figure}[h]
\centering
\includegraphics[width=.8\linewidth]{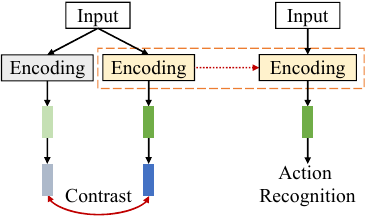}
\caption{
Left: Our proposed contrastive learning framework.
Right: The process of validation in downstream tasks.
} 
\label{fig: fromselftodown}
\end{figure}

\paragraph{From Self-supervised Pre-training to Downstream Tasks}
We first train the encoder in a contrastive manner as described before.
For executing the subsequent downstream tasks, we adopt the process depicted in Figure~\ref{fig: fromselftodown}.
Here we retain the query encoder trained by contrastive learning and employ its output features for the downstream tasks.

\section{Experimental Settings}
\subsection{Datasets} 

\textbf{NTU-60 \& NTU-120.}
NTU-RGB+D~\cite{shahroudy2016ntu} is a multi-view, multi-subject human interaction dataset, and it is also one of the most commonly used datasets in skeleton-related tasks.
Among the samples captured by the dataset, NTU-60 covers 60 action categories and contains more than 56,000 action samples; NTU-120 is expanded to 120 categories and has more than 114,480 samples.
We use the default benchmark of the two datasets for evaluation, namely, cross-view and cross-subject for NTU-60, and cross-set and cross-subject for NTU-120.
We train a model on the data corresponding to some views or subjects, and test on the remaining data.

\noindent\textbf{PKU-MMD I\&II.}
PKU-MMD~\cite{liu2017pku} includes 51 action categories, performed by 66 subjects in three camera views. It contains nearly 20,000 samples.
Referring to \cite{lin2020ms2l}, we conduct experimental verification using the cross-subject evaluation protocol on both PKU-MMD I and II.
Specially, PKU-MMD I consists of 1076 long video sequences spanning the 51 action categories, performed by 66 subjects across three camera views; while PKU-MMD II comprises 2000 short video sequences covering 49 action categories, performed by 13 subjects in three camera views.

\subsection{Implementation Details}

All experiments are executed on NVIDIA GeForce RTX 3090 with Pytorch platform~\cite{paszke2017automatic}.
We adopt SGD with momentum~\cite{qian1999momentum} as the optimiser for both contrastive learning and downstream tasks.
The training paradigm in Moco~\cite{he2020momentum} is borrowed to train our contrastive learning algorithm.
A linear decay strategy for the learning rate is adopted, in which the learning rate is reduced to 1/10 of the previous epoch at certain epoch.
As PKU-MMD II has a very limited number of samples, compared with other datasets, we adopt slightly different training parameters for it.
The implementation details are shown in Table~\ref{table: training}. 

We take action recognition as an example to give the training details of downstream tasks. The training of other tasks is similar, with only minor differences.
For example,
for action retrieval, 
we take the k-nearest neighbours~\cite{cover1967nearest} as the optimisation algorithm. and the number of neighbours used for kNN is set to 1.
For transfer learning and semi-supervised learning, we take exactly the same parameters as for action recognition on NTU dataset, except that the initial learning rate is set to 0.1 and batch size is set to 32.


\section{Additional Experimental Results}

\begin{table}[h]
\centering
\caption{A comparison of different views of the skeleton sequences in terms of x-sub on NTU-60. 'J', 'M' and 'B' represent the Joint-, Motion- and Bone-based inputs.
}
\resizebox{.8\width}{!}{
\begin{tabular}{c c c c c}
\hline
\textbf{Method} & \textbf{J} & \textbf{M} & \textbf{B} & \textbf{J+M+B} \\ \hline
CrosSCLR~\cite{Li_2021_CVPR} &72.9 &72.7 &75.2 &77.8\\
AimCLR~\cite{guo2022contrastive} &74.3 &66.8 &73.2 &78.9 \\
HiCo-GRU~\cite{dong2023hierarchical} &80.6  &78.2 &80.3 & 82.6\\
HiCo-LSTM~\cite{dong2023hierarchical} &81.4 &78.9 &81.0 &83.8 \\
HiCo-Transformer~\cite{dong2023hierarchical}  &81.1 &76.2 &80.3 &83.4 \\
\hline
SCD-Net (Ours) &\textbf{86.6} &\textbf{82.9} &\textbf{82.8} &\textbf{87.3} \\
\hline 
\end{tabular}}
\label{tab: mutiview}
\end{table}

\begin{figure}[!t]
\centering
\subfloat[]{
\includegraphics[scale=0.2]{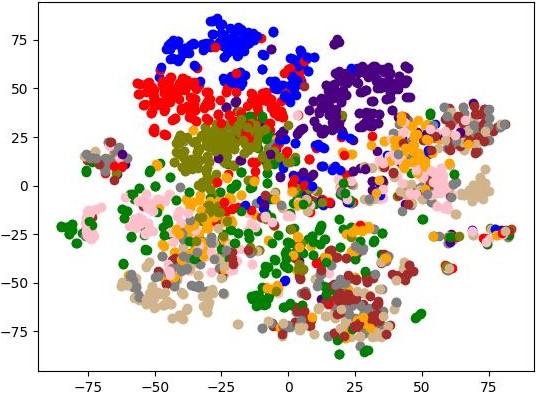}}
\subfloat[]{
\includegraphics[scale=0.2]{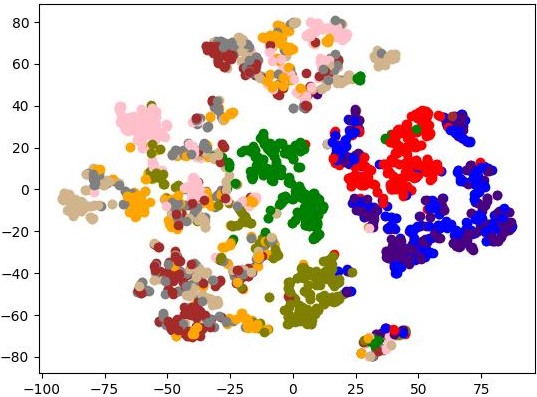}}
\subfloat[]{
\includegraphics[scale=0.2]{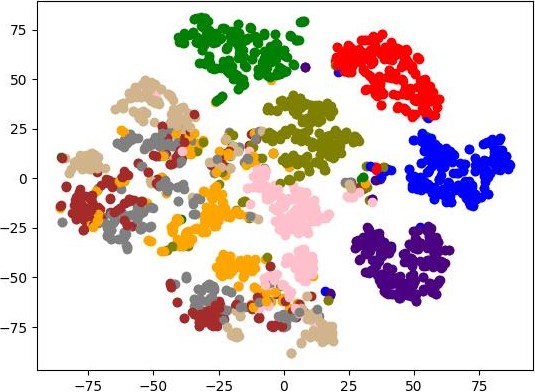}}
\caption{
The t-SNE visualisation results of 10 randomly selected categories.
(a) Original data, DBI: 6.7;
(b) Feature extractor only, DBI: 4.8;
(c) With decoupling, DBI: 3.0.
The Davies-Bouldin Index (DBI) provides a quantitative evaluation of the clustering result (the higher the better).
}
\label{fig: decoupling}
\end{figure}

\paragraph{Different Views}
Our default setting uses joint-based features as input, achieving an accuracy of 86.6\%. We also investigated the performance under different representations following \cite{Li_2021_CVPR}. The results presented in Table~\ref{tab: mutiview} indicate that the corresponding accuracy of motion representation and bone representation are 82.9\% and 82.8\%, respectively. 
Follow \cite{Li_2021_CVPR} for fusion, we also train SCD-Net on multiple views jointly, and adopt linear evaluation on each view of the models, the final performance achieves 87.3\%.
Those results also confirm the potential of our method.

\begin{figure*}[!t]
\centering
\subfloat[Original sequence.]{
\includegraphics[scale=0.95]{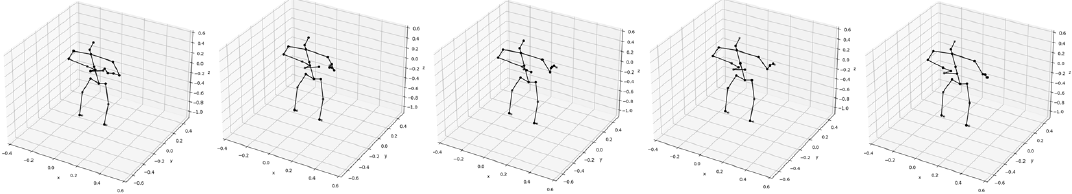}}

\subfloat[Augmented sequence for query encoder with conventional strategy.]{
\includegraphics[scale=0.95]{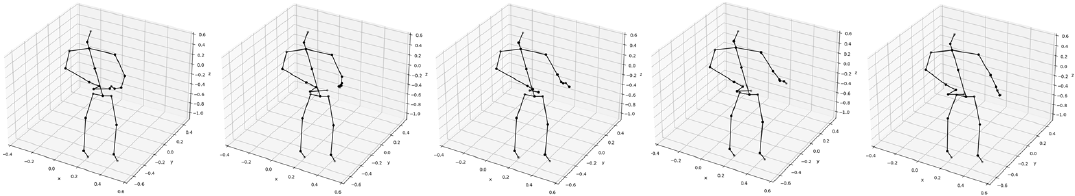}}

\subfloat[Augmented sequence for query encoder with ours strategy.]{
\includegraphics[scale=0.95]{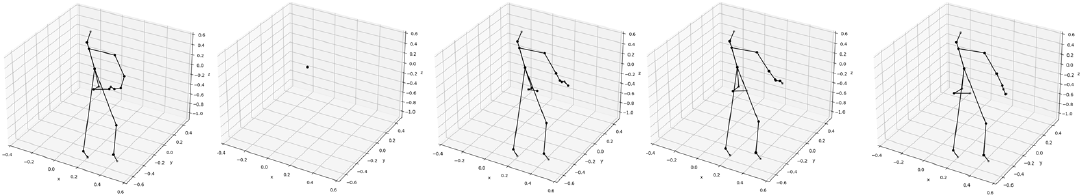}}

\subfloat[Augmented sequence for key encoder with conventional strategy.]{
\includegraphics[scale=0.95]{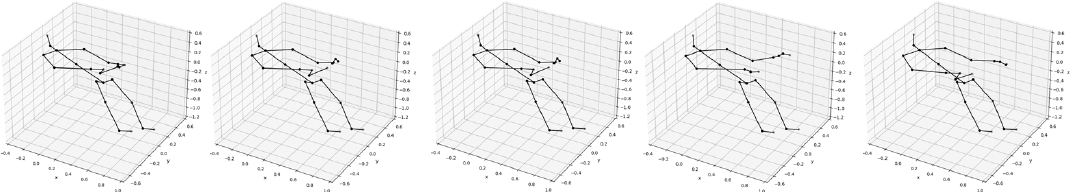}}

\subfloat[Augmented sequence for key encoder with ours strategy.]{
\includegraphics[scale=0.95]{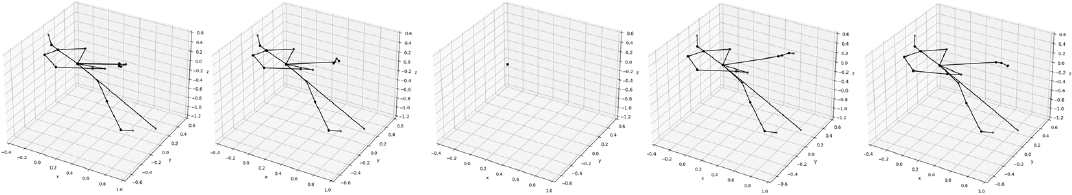}}
\caption{
Visualisation of data augmentation for Query and Key encoders. Five frames uniformly sample from the input sequence are used.
}
\label{fig: vis_data}
\end{figure*}

\paragraph{Visualisation of The Decoupling Encoder}
As shown in Fig.~\ref{fig: decoupling}, we use t-SNE~\cite{van2008visualizing} and DBI~\cite{davies1979cluster} to analyse the representation ability of the features obtained by SCD-Net. 
Fig.~\ref{fig: decoupling}(a) corresponds to the original skeleton sequence without any processing.
Fig.~\ref{fig: decoupling}(b) removes the spatial and temporal decoupling operations from SCD-Net, and only keeps a feature extractor in the encoder. 
Compared with Fig.~\ref{fig: decoupling}(a), the clustering result is much better, demonstrating the effectiveness of feature extractor.
With the decoupling module, Fig.~\ref{fig: decoupling}(c) shows that the samples corresponding to the same category are further aggregated, and the corresponding DBI drops to 3.0. 
The results demonstrate the significance and effectiveness of our decoupling encoder.

\paragraph{Visualisation for Masking}
As shown in Fig.~\ref{fig: vis_data}, 
we visualise the original data and the correspondence augmented data.
It is notable that after conventional augmentation, the relative scale, angle and other indicators of the original skeleton have changed to a certain extent.
However, this change is extremely easy to recover as the basic pattern of the data has not changed.
In contrast, with the structurally constrained masking in the spatial and temporal domains, the structural information of the original sequence is significantly corrupted.
In addition, since we introduce randomness when performing data augmentation, it also makes the input of query encoder and key encoder not absolutely similar, which further promotes the network to learn more robust and discriminative features.

\end{document}